\documentclass[twocolumn,english]{ecai2012}
\pdfoutput=1
\usepackage{mathptmx}
\usepackage[T1]{fontenc}
\usepackage[utf8]{inputenc}
\usepackage{float}
\usepackage{url}
\usepackage{amsthm}
\usepackage{amsmath}
\usepackage{amssymb}
\usepackage{graphicx}
\usepackage{hyperref}

\makeatletter

\newcommand{\noun}[1]{\textsc{#1}}
\providecommand{\tabularnewline}{\\}
\floatstyle{ruled}
\newfloat{algorithm}{tbp}{loa}
\providecommand{\algorithmname}{Algorithm}
\floatname{algorithm}{\protect\algorithmname}

\usepackage{babel}
\usepackage{algorithmicx}
\usepackage{algpseudocode}
\usepackage{xcolor}
\usepackage[T1]{fontenc}

\@ifundefined{showcaptionsetup}{}{%
 \PassOptionsToPackage{caption=false}{subfig}}
\usepackage{subfig}
\makeatother

\usepackage{babel}
\begin{document}

\title{Asynchronous Decentralized Algorithm for Space-Time Cooperative Pathfinding}

\author{Michal Čáp\institute{FEL, \emph{Czech Technical University in Prague},
Czech Republic, e-mail: \sf \{michal.cap, jiri.vokrinek, michal.pechoucek\}@agents.fel.cvut.cz}
\and Peter Novák\institute{EEMCS, \emph{Delft University of Technology},
The Netherlands, e-mail: \sf P.Novak@tudelft.nl} \and Jiří Vokřínek\footnotemark[1]
\and Michal Pěchouček\footnotemark[1]}
\maketitle
\begin{abstract}
Cooperative pathfinding is a multi-agent path planning problem where
a group of vehicles searches for a corresponding set of non-conflicting
space-time trajectories. Many of the practical methods for centralized
solving of cooperative pathfinding problems are based on the prioritized
planning strategy. However, in some domains (e.g., multi-robot teams
of unmanned aerial vehicles, autonomous underwater vehicles, or unmanned
ground vehicles) a decentralized approach may be more desirable than
a centralized one due to communication limitations imposed by the
domain and/or privacy concerns.

In this paper we present an asynchronous decentralized variant of
prioritized planning ADPP and its interruptible version IADPP. The
algorithm exploits the inherent parallelism of distributed systems
and allows for a speed up of the computation process. Unlike the synchronized
planning approaches, the algorithm allows an agent to react to updates
about other agents' paths \emph{immediately} and invoke its local
spatio-temporal path planner to find the best trajectory, as response
to the other agents' choices. We provide a proof of correctness of
the algorithms and experimentally evaluate them on synthetic domains.
\end{abstract}

\section{Introduction \label{sec:Introduction}}

When mobile agents operate in a shared space, one of the essential
tasks for them is to prevent collisions among themselves, possibly
even to maintain a safe distance from each other. Prominent examples
of domains requiring robust collision avoidance techniques are different
kinds of autonomous multi-robotic systems, next-generation air traffic
management systems, road traffic management systems etc.

A range of methods is being currently employed to realize a safe operation
of agents within a shared space. Some of the methods assume a cooperative
setting where all the involved agents work together to solve their
mutual conflicts, others assume a non-cooperative setting where the
agents cannot coordinate their actions, and yet others consider pursuit-evasion
adversarial scenarios where a solution is a trajectory that is collision
free against the worst-case behaviour of other agents. In this work,
we focus on the cooperative pathfinding.

Cooperative path planners are used to plan the routes for a number
of agents, taking in consideration the objectives of each agent while
avoiding conflicts between the agents' paths. If the agents execute
the resulting multi-agent plan precisely, it is guaranteed that the
agents will not collide. Centralized solvers in literature are based
either on global search or decoupled planning. Global search methods
find optimal solutions, but they do not scale well for higher (over
ten) numbers of conflicting agents. One of the most efficient optimal
solvers for cooperative pathfinding on grids has been introduced by
Standley in 2010~\cite{Standley10}.

Decoupled approaches are incomplete, but can be fast enough for real-time
applications e.g., in the video-game industry. One of the the standard
technique employed in gaming is the \emph{Local Repair A{*}} (LCA{*})
algorithm~\cite{Silver05}. In LCA{*} each agent plans a path independently
and tries to follow it to the goal position. If a collision occurs
during the path plan execution, the agent replans the remainder of
the route from the collision position taking into account the positions
in its vicinity occupied by the other agents involved in the collision.
Due to its greedy and reactive nature, the method does not perform
well in cluttered environments with bottlenecks and can generate cycles,
or otherwise aesthetically unpleasant, or inefficient behaviours of
agents~\cite{Pottinger99}. To mitigate these problems, Silver~\cite{Silver05}
introduced \emph{Cooperative A{*}} (CA) a cooperative pathfinding
algorithm based on the idea of prioritized planning~\cite{Erdmann87onmultiple}. 

In prioritized planing, each agent is assigned a priority and the
planning process proceeds sequentially agent after agent in the order
of the agents' priorities. The first agent plans its path using a
single-agent planner disregarding the positions and objectives of
other agents. Each subsequent agent models the paths of the higher-priority
agents as moving obstacles and plans its path such that the collisions
with the higher-priority agents' paths are avoided. Such an approach
has been shown to be effective in practice~\cite{Ferrari1998219}.
The quality of the generated solution is sensitive to the assigned
priority ordering, however there is a simple heuristic for choosing
an efficient ordering for the prioritized planning~\cite{BergO05}.

Recently, Velagapudi presented a decentralized prioritized planning
technique for large teams of mobile robots~\cite{VelagapudiSS10}.
The method is shown to generate the same results as the centralized
planner. However, the formulation of the decentralized algorithm is
based on the assumption that the robots have a \textquotedbl{}distributed
synchronization mechanism allowing them to wait for all team mates
to reach a certain point in algorithm execution\textquotedbl{}~\cite{VelagapudiSS10}
and thus it does not exploit the asynchrony common in distributed
systems. Rather the computation proceeds in iterations and the agents
wait for each other at the end of each algorithm iteration. As a consequence,
the algorithm does proceeds in synchronous rounds, where the length
of a round is dictated by the agent performing the longest computation
due to either a high workload, or low computational resources available.

After stating the cooperative pathfinding problem and exposing the
underlying ideas of the state-of-the-art prioritized planning approaches
in Sections~\ref{sec:Cooperative-Pathfinding-Problem} and~\ref{sec:Prioritized-Planning},
Section~\ref{sec:Asynchronous-Prioritized-Plannin} presents the
main contribution of the paper, the \emph{asynchronous decentralized
prioritized planning algorithm} (ADPP). ADPP, is an extension of the
synchronized decentralized prioritized planning algorithm (SDPP),
which removes the assumptions of synchronous execution of the decentralized
algorithm. Besides the generic form of the ADPP algorithm, we also
present a locally asynchronous modification of the ADPP algorithm
(\emph{interruptible ADPP}, IADPP) enabling interruptible path planner
execution so that the individual agents can react to updates received
from their peers more swiftly. To prove termination and correctness
properties of ADPP and IADPP, we provide a new proof of termination
and correctness also for the SDPP algorithm. Our proof is an alternative
to the original argument presented in~\cite{VelagapudiSS10}. We
implemented and extensively evaluated the discussed algorithms on
a number of synthetic scenarios. Section~\ref{sec:Evaluation} provides
both an illustrative theoretical comparison of the SDPP and ADPP approaches,
as well as details the experimental evaluation of the introduced algorithms.
The experimental validations show that the asynchronous versions of
the prioritized planning algorithm offer better runtime performance,
as well as improved use of the available computational resources.

\section{Cooperative Pathfinding Problem \label{sec:Cooperative-Pathfinding-Problem}}

Consider $n$ agents $a_{1},\ldots,a_{n}$ operating in an Euclidean
space $\mathcal{W}$. Each agent $a_{i}$ is characterized by its
starting and goal positions $start_{i},\, dest_{i}$ respectively.
The task is to find a set of space-time trajectories $P=\{p_{1},\ldots,p_{n}\}$,
such that $p_{i}:\mathbb{R}\rightarrow\mathcal{W}$ is a mapping from
time points to positions in $\mathcal{W}$, $p_{i}(0)=start_{i}$,
$p_{i}(t_{i})=dest_{i}$ and the trajectories are mutually collision
free, i.e., $\forall i,j:\, i\neq j\Rightarrow\neg C(p_{i},p_{j})$,
where $C(p_{i},p_{j})$ denotes a space-time mutual collision relation
between $p_{i}$ and $p_{j}$. Informally, two trajectories collide
(are in a conflict) when the trajectories touch, or intersect. That
is $p_{i}[t^{\prime}]=p_{j}[t^{\prime}]$ for some timepoint $t^{\prime}$.
However, more complex collision relations can be considered, such
as those considering a minimal separation range between trajectories,
etc. $t_{i}^{\mathit{dest}}=\min\{t_{i}\mid p_{i}[t_{i}]=g_{i}\}$
denotes the shortest timepoint in which the agent $a_{i}$ reaches
its destination $dest_{i}$. As a solution quality metric we use the
cumulative time spent by agents navigating their trajectories defined
as $\mathit{dur}(P)=\sum_{i=1}^{n}t_{i}^{\mathit{dest}}$. The cost
of solution $P$ is defined as $cost(P)=\frac{dur(P)-dur(P')}{dur(P')}$,
where $P'$ is the set of best trajectories for each agent if the
collisions are ignored.

\section{Prioritized Planning \label{sec:Prioritized-Planning}}

In general, the complexity of complete approaches to multi-agent path
planning grows exponentially with the number of agents. Therefore,
the complete approaches often do not scale-up well and hence are often
not applicable for nontrivial domains with many agents. To plan paths
for a high number of agents in a complex environment, one has to resort
to one of the incomplete, but fast approaches. A simple method often
used in practice is \emph{prioritized} planning \cite{Erdmann87onmultiple,BergO05,Bennewitz02Planning}.
In prioritized planning the agents are assigned a unique priority.
In its simplest form, the algorithm proceeds sequentially and agents
plan individually from the highest priority agent to the lowest one.
The agents consider the trajectories of higher priority agents as
constraints (moving obstacles), which they need to avoid. It is straightforward
to see that when the algorithm finishes, each agent is assigned a
trajectory not colliding with either higher priority agents, since
the agent avoided a collision with those, nor with lower priority
agents who avoided a conflict with the given trajectory themselves.

\begin{figure}
\begin{centering}
\includegraphics[width=4cm]{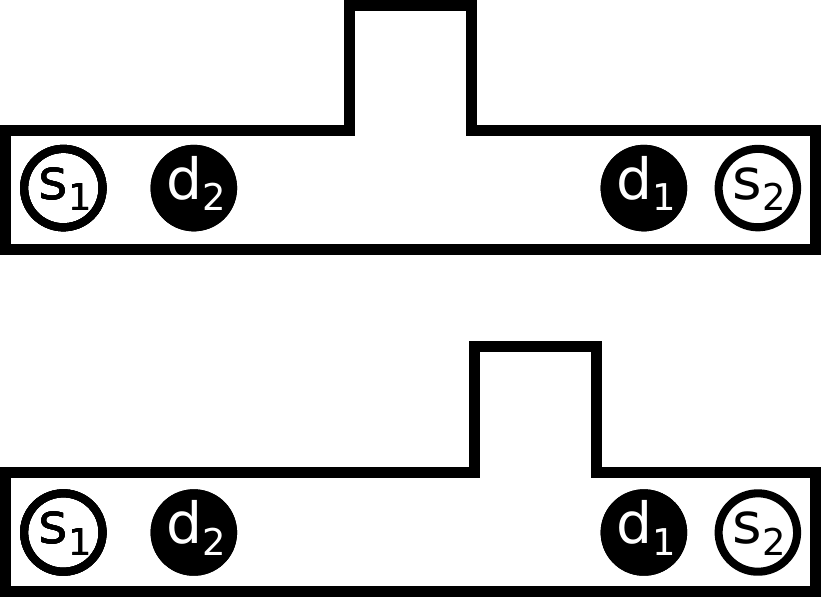}
\par\end{centering}

\caption{\label{fig:Top: Hard_problems_for_prioritized_planner}Top: example
of a problem to which a prioritized planner will not find a solution.
The first agent plans its optimal path first, but such a trajectory
is in conflict with all feasible trajectories of the second agent.
Bottom: example of a problem to which a prioritized planner will find
a solution only if agent 1 has a higher priority than agent 2. }
\end{figure}

The complexity of the generic algorithm grows linearly with the number
of agents, which makes the approach applicable for problems involving
many agents. Clearly, the algorithm is greedy and incomplete in the
sense that agents are satisfied with the first trajectory not colliding
with higher priority agents and if a single agent is unable to find
a collision-free path for itself, the overall path finding algorithm
fails. The benefit, however, is fast runtime in relatively uncluttered
environments, which is often the case in multi-robotic applications.
Prioritized planner is also sensitive to the initial prioritization
of the agents. Both phenomena are illustrated in Figure~\ref{fig:Top: Hard_problems_for_prioritized_planner}
that shows a simple scenario with two agents desiring to move from
$s_{1}$ to $d_{1}$ ($s_{2}$ to $d_{2}$ resp.) in a corridor that
is only slightly wider than a single agent. The scenario assumes that
both agents have identical maximum speeds.

\subsection{Computing best response}

During prioritized planning, an individual agent searches the shortest
path to its destination considering other higher-priority agents as
moving obstacles during the planning process. Ideally, the agent should
compute the best possible trajectory, a best response to the trajectories
of the higher-priority agents. To find such a best response, the agent
needs to solve a motion planning problem with dynamic obstacles, which
is a significantly more complex task than the motion planning with
static obstacles since a new independent time dimension has to be
considered during planning. Henceforth, we will denote the single-agent
best-response planer process as a function $\textrm{\textnormal{\textsc{Best-path}}}_{i}(\mathit{start},\mathit{dest},\mathit{avoids})$,
which returns the selected best trajectory for the agent $i$, starting
in the position $\mathit{start}$, eventually reaching the position
$\mathit{dest}$ and at the same time not colliding with any of the
trajectories in the set $\mathit{avoids}$. Note, we do not precisely
specify what the best trajectory means, the notion can be application-specific
for the individual agent. For simplicity, however, in the following
we assume the notion of the best path to correlate with time-optimality
of trajectories, i.e., the how fast a given agent can navigate along
the trajectory given  its specific motion dynamic constraints.

\subsection{Centralized Algorithm}

A collision-free operation of a multi-agent team can be ensured by
forcing all agents to communicate their objectives to a centralized
planner, which centrally computes a solution and informs the agents
about the trajectory they have to follow in order to maintain the
conflict-free operation. As a baseline for evaluation of performance
of the latter introduced algorithms, we use the \emph{cooperative
A{*} algorithm}~\cite{Silver05}. Cooperative A{*} is a centralized
algorithm for cooperative path finding based on prioritized planning
employing the well-known A{*} trajectory planning algorithm on grids~\cite{Hart_Astar}.
Algorithm~\ref{alg:CA} lists the pseudocode of the cooperative A{*}
algorithm. We discussed the correctness of this generic algorithm
above.

\global\long\def\start{\mathit{Start}_{i}}
\global\long\def\dest{\mathit{Dest}_{i}}
\global\long\def\priority{I}
\global\long\def\nagents{\mathit{N}}

\global\long\def\currpath{\mathit{Path_{i}}}
\global\long\def\avoids{\mathit{Agentview}_{i}}
\global\long\def\inbox{\mathit{Mailbox}_{i}}

\algrenewcommand{\algorithmiccomment}[1]{{\color{gray} $\triangleright$ \emph{#1} $\triangleleft$}}

\algblockdefx[MsgHandler]{MsgHandler}{EndMsgHandler}
[2]{\textbf{message handler} \noun{#1}(#2)}
{\textbf{end message handler}}

\begin{algorithm}
\caption{\label{alg:CA} Centralized Prioritized Planning (Cooperative A{*})}

\begin{algorithmic}[1]

\Ensure After the algorithm finishes, $\currpath$ contains the final
computed path for the agent with priority $i$. If the agent couldn't
find a path not colliding with higher priority agents, $\currpath$
stores $\emptyset$.

\Statex

\Procedure{CA}{$\langle\mathit{start}_{1},\mathit{dest}_{1}\rangle,\ldots,\langle\mathit{start}_{n},\mathit{dest}_{n}\rangle$}

	\State $\mathit{Avoids}\gets\emptyset$

	\For{$i\gets1\ldots n$}

		\State$\currpath\gets$\Call{Best-path$_i$}{$\mathit{start}_{i},\mathit{dest}_{i},\mathit{Avoids}$}

		\State $\mathit{Avoids}\gets\mathit{Avoids}\cup\{\currpath\}$

	\EndFor

\EndProcedure

\Statex

\Function{Best-path$_i$}{$\mathit{start},\mathit{dest},\mathit{avoids}$}

	\State{\textbf{return} the best $\mathit{path}$ from $\mathit{start}$
to $\mathit{dest}$ not conflicting with}

	\State{\hskip9.5mm{}any of the paths in $\mathit{avoids}$. Otherwise
return $\emptyset$.}

\EndFunction

\end{algorithmic}
\end{algorithm}

\subsection{Decentralized Algorithms}

A decentralized algorithm for solving cooperative pathfinding problems
by means of prioritized planning has been presented in \cite{VelagapudiSS10}.
The algorithm is synchronous in that it contains synchronization points
in the program execution through which all agents proceed simultaneously.
Due to the synchronous nature of the algorithm, we will refer to this
algorithm as \emph{synchronized decentralized prioritized planning}
(SDPP). Algorithm~\ref{alg:SDPP} lists the pseudocode of SDPP. We
slightly adapted the algorithm listing for exposition purposes and
comparison with the later introduced algorithms. Note that in the
decentralized setting we assume communication to be reliable and the
communication channel preserves the order of messages they were sent
in. Furthermore, the algorithm assumes that before the start of the
algorithm, each agent is assigned a unique priority, an ordinal $\priority\in1\ldots\nagents$,
where $\nagents$ is the number of agents taking part on the algorithm
run (the lowest $\priority$ means the highest priority). The algorithm
is also locally asynchronous and we assume safe (thread-safe) access
to global variables (denoted by capitalized identifiers). To simplify
the exposure, the thread-barrier locking mechanism is omitted from
the pseudocode.

\begin{algorithm}
\caption{\label{alg:SDPP} Synchronized Decentralized Prioritized Planning\protect \\
\Comment{pseudocode for the agent $i$}}

\begin{algorithmic}[1]

\Ensure After the algorithm finishes, $\currpath$ contains the final
computed path. If no solution was found, $\currpath$ stores $\emptyset$.

\Statex

\Procedure{SDPP}{$\mathit{start},\mathit{dest},\mathit{nagents,}priority$}

	\State {\footnotesize $\start\gets start$; $\dest\gets dest$}{\footnotesize \par}

	\State {\footnotesize $\nagents\gets nagents$; $I\gets\mathit{priority}$}{\footnotesize \par}

	\State {\footnotesize $\avoids\gets\emptyset$; $\currpath\gets\emptyset$}{\footnotesize \par}

	\Repeat

		\State\Call{Check-consistency-and-plan}{}

		\State \textbf{wait} for all other agents to finish the planning
iteration

	\Until{global termination detected}

\EndProcedure

\Statex

\Procedure{Check-consistency-and-plan}{}

		\If{$\currpath\textrm{ collides with }\avoids$}

			\State\Comment{Work on a copy of the $\avoids$}

			\State $\currpath\gets$\Call{Best-path$_i$}{$\start,\mbox{\ensuremath{\dest}},\avoids$}

			\ForAll{$j\gets I+1\ldots N$}

				\State \Call{Send-inform-to-$j$}{$\priority,\currpath$}

			\EndFor

		\EndIf

\EndProcedure\Statex

\MsgHandler{Receive-inform}{$j,path$}

	\State $\avoids\gets(\avoids\setminus\langle j,\_\rangle)\cup\{\langle j,\mathit{path}\rangle\}$

\EndMsgHandler

\end{algorithmic}
\end{algorithm}

The algorithm proceeds in iterations. In each iteration the agents
compute the best path if necessary and subsequently communicate it
to the lower priority agents. An agent must recompute its trajectory
in the case its current path collides with some trajectories of higher
priority agents computed and communicated in the previous iterations.
Upon receiving an \noun{inform} message, the agent simply replaces
the information about the trajectory of the sender agent in its $\avoids$
set. Note, the algorithm is asynchronous, therefore the trajectory
planning routine $\textrm{\textsc{Best-path}}_{i}$ operates on a
copy the $\avoids$ set.

The algorithm finishes when all the agents cease to communicate and
either hold a trajectory, or they were not able to find a collision-free
trajectory. We assume that the global termination condition is detected
by some concurrently running global state detection algorithm, such
as the Chandy and Lamport's snapshot algorithm~\cite{Chandy_snapshot}.

The presented SDPP algorithm is correct in that when it finds a solution
for all the participating agents, the paths are mutually collision
free. However, the algorithm is incomplete in the sense that there
are situations when the algorithm fails to find a solution for all
the participating agents, even though such a solution exists. In order
to facilitate and simplify exposition of the later introduced algorithms,
we developed a new alternative proof of the SDPP algorithm, which
deviates from the original one devised by the authors of SDPP~\cite{VelagapudiSS10}. 

To see the correctness of the SDPP algorithm we need to show that
firstly, the algorithm terminates, and secondly that the resulting
paths are mutually collision free. 
\begin{proof}[Proof (SDPP termination):]
 First of all, we need to show that the algorithm finishes. That
is, each agent $i$ eventually stops sending \noun{inform} messages.
We proceed by induction on the individual agent priority $i$. 
\begin{description}
\item [{initial~step}] since there is no agent with priority higher than
agent $a_{1}$, the highest priority agent $a_{1}$ informs the lower
priority agents only once in the first iteration of the algorithm
and from then on it remains silent since its path will always be non-colliding
with an empty set of paths - there are no higher priority agents to
inform this agent about an update of the situation.
\item [{induction~step}] Let's assume the following induction hypothesis:
\emph{``after the agents with priorities $1\ldots k-1$ stopped communicating,
eventually also the agent with priority $k$ stops sending }\emph{\noun{inform}}\emph{
messages''}. Let's assume this is not the case and there is a situation
such that the agent $k$ would end up sending \noun{inform} messages
forever. For such to occur, the agent however must have its mailbox
continually being filled with \noun{inform} messages so that it's
\noun{Receive-inform} handler routine gets invoked infinitely many
times. In a consequence the agent would possibly need to recompute
its best path and subsequently inform the lower priority agents infinitely
often. That however implies existence of a sender for each such a
message and hence by necessity there must be at least one agent with
priority higher than $k$ which keeps sending \noun{inform} messages
forever, which contradicts the induction hypothesis.
\end{description}
\end{proof}
As a consequence of the consecutive silencing of agents from high
to lower priorities, it's also relatively straightforward to see that
the SDPP algorithm makes at most $\nagents$ iterations before it
terminates.

Note, that not necessarily it is the agent with the lowest priority
which stops communicating the last. In the case a lower priority agent
computes a route which is not in a conflict with a current set of
temporary routes of the higher priority agents, nor with any routes
they will compute later on, its reactions to receiving \noun{inform}
messages will be silent and won't result in further cascade of communication.

\begin{proof}[Proof (SDPP correctness):]
 To see that after the algorithm termination the variables $\currpath$
store a set of non-conflicting paths is rather straightforward. Since
each agent eventually sends its last \noun{inform} message and cedes
to communicate, each agent with priority lower than its own eventually
collects all the last \noun{inform} messages from all the higher priority
agents, together with their ultimate paths (being either a valid path,
or $\emptyset$). At that moment, all the couples $\langle j,\mathit{Path}_{j}\rangle$
for all $j>i$ are stored in the set $\avoids$ of the agent with
priority $i$. Subsequently the agent eventually invokes the \noun{Check-consistency-and-plan}
routine for the last time and thus either $\currpath$ will end up
unchanged, recomputed and again non-conflicting with either of $\langle j,\mathit{Path}_{j}\rangle$
for all $j<i$, or being invalid ($\emptyset$). Finally, the agent
informs all the agents with priorities lower than $i$ and cedes to
communicate. At the moment when the last agent stops communicating,
all the $\mathit{Path}_{i}$ variables are either set correctly, or
the algorithm failed to find a solution for some of the participating
agents.
\end{proof}
As we already noted above, the SDPP algorithm is incomplete. To see
that, consider a situation in which the agent with the highest priority
makes a choice which later on constraints some of the lower priority
agents so that they are unable to find a solution. In the case there
would be a locally worse choice for the highest priority agent, which
however would enable the lower priority agents to find valid solutions,
the SDPP algorithm does not facilitate re-consideration of the first
choice, nor some backtracking mechanism.

During the algorithm computation, it can however happen that an agent
$i$ sets its $\mathit{Path}_{i}$ to $\emptyset$ and later reconsiders
this decision. This happens when among paths of the higher priority
agents there are conflicting couples, but those agents did not manage
to resolve the collisions yet and at the same time the lower priority
agent $i$ is temporarily not able to route around the space occupied
by the temporary paths of the higher priority agents.

Note that in the distributed prioritized planning, one can use a simple
marking-based termination-detection mechanism. Following the proof
of termination, agent $i$ can mark its path \emph{final} if the path
of agent priority $i-1$ in $\avoids$ is marked final. The initial
path of $a_{1}$ is final. When an agent sends his final path to a
lower-priority agent, the higher-priority agent can safely terminate
its computation. When the final path is generated by the lowest-priority
agent, the computation terminated globally.

\section{Asynchronous Prioritized Planning \label{sec:Asynchronous-Prioritized-Plannin}}

The SDPP algorithm does not fully exploit the parallelism of the distributed
system, a drawback stemming from its synchronous nature. The running
time of a single iteration of the SDPP algorithm is largely influenced
by the speed of the computationally slowest agent of the group. In
every iteration, the agents which finished their trajectory planning
routine faster, or did not have to re-plan at all sit idle while waiting
for the agents with higher workload in that iteration (or simply slower
computation), even though they could theoretically resolve some of
the conflicts they have among themselves in the meantime and thus
speed up the overall algorithm run. 

To improve the performance of the decentralized cooperative path finding,
we propose an \emph{asynchronous decentralized prioritized planning
algorithm} (ADPP), an asynchronous variant of SDPP. Algorithm~\ref{alg:ADPP}
lists the pseudocode of ADPP.

\begin{algorithm}
\caption{\label{alg:ADPP} Asynchronous Decentralized Prioritized Planning\protect \\
\Comment{pseudocode for the agent $i$}}

\begin{algorithmic}[1]

\Procedure{ADPP}{$\mathit{start},\mathit{dest},\mathit{nagents},\mathit{priority}$}

	\State {\footnotesize $\start\gets start$; $\dest\gets dest$}{\footnotesize \par}

	\State {\footnotesize $\nagents\gets nagents$; $I\gets\mathit{priority}$}{\footnotesize \par}

	\State {\footnotesize $\avoids\gets\emptyset$; $\currpath\gets\emptyset$}{\footnotesize \par}

	\Repeat

		\State $\mathit{CheckFlag}_{i}\gets\mathit{false}$

		\State \Call{Check-consistency-and-plan}{}

		\State \textbf{wait} for $\mathit{CheckFlag}_{i}$ , or global
termination

	\Until global termination detected

\EndProcedure

\Statex

\MsgHandler{Receive-inform}{$j,path$}

	\State $\avoids\gets(\avoids\setminus\langle j,\_\rangle)\cup\{\langle j,\mathit{path}\rangle\}$

	\State $\mathit{CheckFlag}_{i}\gets\mathit{true}$

\EndMsgHandler

\end{algorithmic}
\end{algorithm}

The main deviation from the SDPP listed in Algorithm~\ref{alg:SDPP}
is the formulation of the waiting condition in the main loop of the
algorithm. While each agent of the group waits for all the other to
finish in the SDPP algorithm, in the ADPP algorithm, they break their
idle upon receiving the next \noun{inform} message or a need to process
updated $\avoids$, in the case the agent received a number of \noun{inform}
messages during the time it was occupied with planning its own trajectory.
The arrival of a new \noun{inform} message and thus the need to re-check
the consistency of the currently computed path with respect to the
new information is indicated by the state of the $\mathit{CheckFlag}_{i}$
variable.

The proof of correctness of the ADPP algorithm follows exactly the
correctness proof of the SDPP algorithm above. Note, in the SDPP proof,
the condition that the algorithm proceeds in a synchronized manner
was never exploited. The ADPP algorithm terminates for exactly the
same reasons as SDPP. Namely, the agent with the highest priority
stops communicating right after it computes its path for the first
time and in consequence the agents with lower priority consecutively
cede to communicate later on as well until the algorithm terminates.
 The argument for ADPP incompleteness follows the incompleteness
argument for SDPP as well.

\subsection*{Interruptible ADPP}

The ADPP algorithm exploits the potential speed up with respect to
the inter-agent communication. However, while the agent is computing
the best path in the current situation, messages keep arriving. In
a consequence, it can happen that an individual agent's computation
returns from the path planning routine \noun{$\textrm{\textsc{Best-path}}_{i}$}
only to find out that large part of the work was invalidated by some
later received messages. This reveals a potential further speed-up
of the ADPP algorithm by interrupting the path planning upon reception
of every \noun{inform} message and re-considering the computation
in the light of the newly received message. Algorithm~\ref{alg:IADPP}
lists a pseudocode of a modified ADPP algorithm which pro-actively
interrupts the trajectory planning computation upon receiving every
new \noun{inform} message. Alternatively, it is conceivable to exploit
algorithms for dynamic trajectory planning, which allow topological
changes during the planning process.

\begin{algorithm}
\caption{\label{alg:IADPP} Interruptible Asynchronous Decentralized Prioritized
Planning - pseudocode for the agent $i$}

\begin{algorithmic}[1]

\Procedure{IADPP}{$\mathit{start},\mathit{dest},\mathit{nagents},\mathit{priority}$}

	\State {\footnotesize $\start\gets start$; $\dest\gets dest$}{\footnotesize \par}

	\State {\footnotesize $\nagents\gets nagents$; $\priority\gets\mathit{priority}$}{\footnotesize \par}

	\State {\footnotesize $\avoids\gets\emptyset$; $\currpath\gets\emptyset$}{\footnotesize \par}

	\State \Call{Check-consistency-and-plan}{}

	\State{\textbf{wait} for global termination}

\EndProcedure

\Statex

\MsgHandler{Receive-inform}{$j,path$}

	\State $\avoids\gets(\avoids\setminus\langle j,\_\rangle)\cup\{\langle j,\mathit{path}\rangle\}$

	\State \textbf{asynchronously launch/restart \{}

	\Statex\hspace{\algorithmicindent}\hspace{\algorithmicindent}\Call{Check-consistency-and-plan}{}\textbf{\}}

\EndMsgHandler

\end{algorithmic}
\end{algorithm}

Note, the main \textbf{repeat-until} loop was replaced by simple wait
for the algorithm termination. The repeated consistency check (calls
of the \noun{Check-consistency-and-plan} routine) is secured by its
asynchronous invocation from the \noun{Receive-inform} routine. That
is, the routine is executed in a newly created computation run (thread)
and the call does not wait for its termination, it runs in parallel
to the \noun{Receive-inform} routine from then on. In the case there
is already a concurrent invocation of the \noun{Check-consistency-and-plan}
routine running, it is killed and run anew (restarted) with the updated
$\avoids$ set.

The termination and correctness of the IADPP algorithm stems from
the termination and correctness of the ADPP algorithm. The same proof
applies, since the IADPP modification was strictly local, not affecting
the communication patterns between the participating agents.

\section{Evaluation \label{sec:Evaluation}}

The motivation for introducing the decentralized algorithm and its
asynchronous variants is oriented mainly to the runtime improvements
of the algorithm. Clearly, such a potential improvement is greatly
influenced by the topology of the problem and the selection of agent
priorities. In this section, we first discuss the noticeable features
of the presented algorithms and our expectation on their performance.
Then, we will present experimental evaluations using superconflict
and randomly generated scenarios.

\subsection{Theoretical analysis}

As indicated above, the decentralized approaches should benefit from
the concurrent execution on a higher number of processors (i.e., equal,
or higher than the number of agents). The wall-clock runtime of the
algorithms is expected to be lower for decentralized algorithms, but
there might exist some problem configurations that yield directly
opposite results. In this section we sketch a theoretical analysis
of the impact of the parallelism and asynchronicity and show examples
to demonstrate the presented ideas.

\begin{figure}
\begin{centering}
\includegraphics[width=5cm]{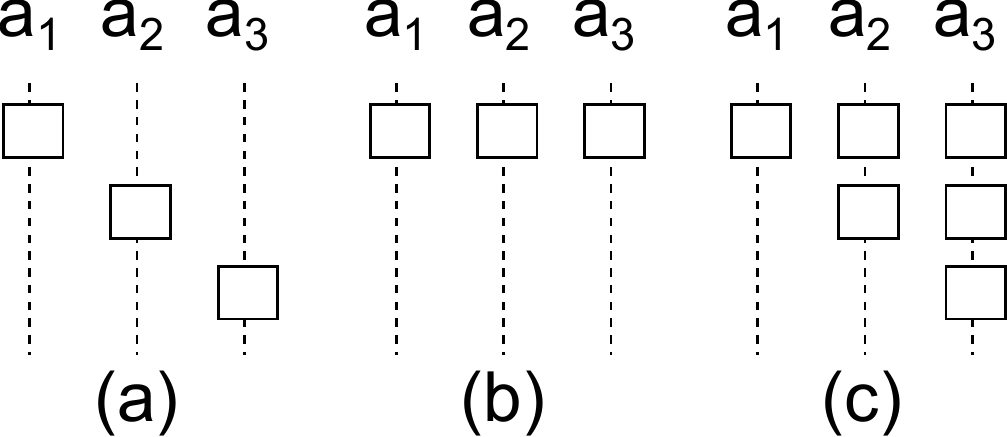}
\par\end{centering}

\caption{\label{fig:execution_c_vs_d}Example of the path search execution
sequence for (a) centralized algorithm, (b) decentralized algorithm
for non-colliding trajectories problem and (c) decentralized algorithm
for mutually-colliding trajectories problem. The boxes represent invocations
of best-response planners.}
\end{figure}

Let us first discuss the differences between the centralized and decentralized
approaches. For simplicity, let the processing time of the best-path
search routine be one time unit for each path searched (one path for
one agent). Figure~\ref{fig:execution_c_vs_d} illustrates an example
of the algorithm execution sequence for three agents, where priorities
of the agents are given from left to right and match the agent indices.
The centralized algorithm simply computes the agents' paths sequentially
in the order of agents' priorities. The total wall-clock runtime is
3 time units here. 

To analyze the algorithm runs in decentralized scenarios, consider
a scenario where the agents have non-conflicting trajectories and
a superconflict scenario, in which the best trajectories of all the
agents collide. In a distributed setting, we assume three parallel
processors, i.e. one for each agent. In the case of non-conflicting
trajectories the agents should be able to fully utilize the inherent
parallelism of the distributed system, so that the wall-clock runtime
of the algorithm is only one time unit. However, in the case of superconflict
scenario the situation is different. Each lower-priority agent has
to recompute his path when a higher-priority agent produces a new
solution. Clearly, the parallel execution has no speed-up effect here
since the wall-clock runtime stays 3 time units. This example provides
an intuition for the bounds of the decentralized algorithm execution
time. One would expect that the wall-clock runtime of a decentralized
algorithm will be equal or lower than the execution time of the centralized
algorithm depending on the scale of coupling between the agents. That
is, informally, on the size of a cluster in which agents' trajectories
influence each other. 

\begin{figure}
\centering{}\subfloat{\centering{}\includegraphics[width=5cm]{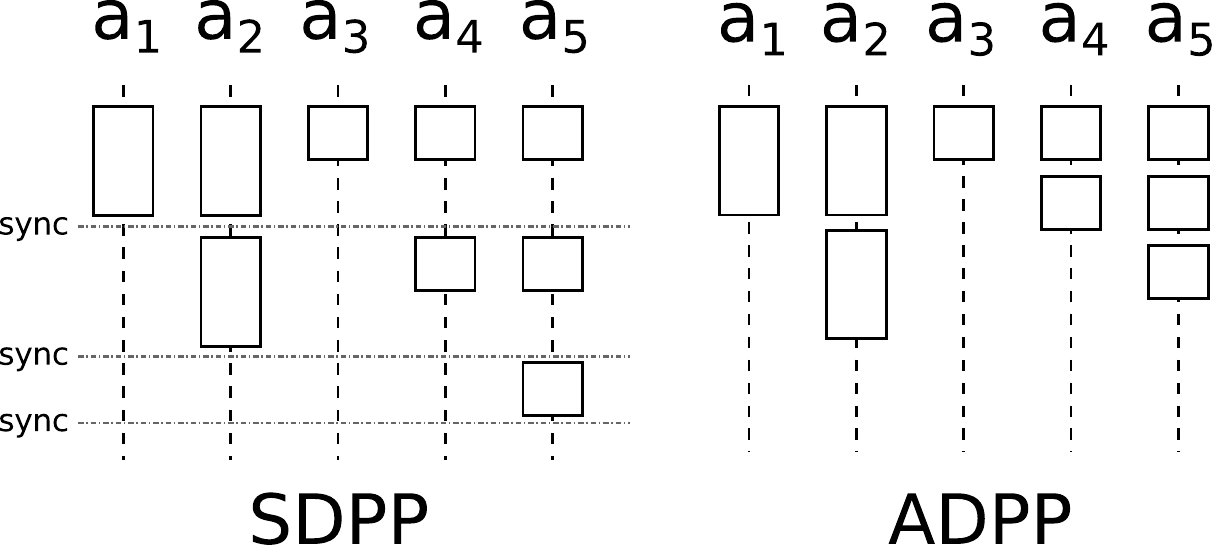}}\caption{\label{fig:Different_plan_length_scenario-1}Sequence diagram showing
the execution of SDPP resolution process and ADPP resolution process
for a scenario with two independent conflict clusters, where agents
in $\{a_{1},a_{2}\}$ and $\{a_{3},a_{4},a_{5}\}$ need different
amount of time to find their best response. }
\end{figure}

However, the situation changes if we assume non-uniform runtimes of
the agents' best-response planers. In such a situation, SDPP may suffer
from significant synchronization overheads. Figure~\ref{fig:Different_plan_length_scenario-1}
illustrates the difference between the synchronous and the asynchronous
variant of the decentralized approach. In this example ADPP exploits
existence of independent conflict clusters and is able to lower the
total wall-clock runtime from 5 to 4 time units. Main distinguishing
feature of the ADPP algorithm over SDPP is that in ADPP an agent starts
resolving conflicts immediately after the agent detects them, while
in SDPP the conflicts are resolved in the next iteration of the algorithm.
Since the duration of one SDPP iteration is determined by the slowest
computing agent, the computational power of faster computing agents
may stay unutilized. This example illustrates how can be the wall-clock
runtime reduced by the asynchronous algorithm.

\begin{figure}
\begin{centering}
\includegraphics[width=5cm]{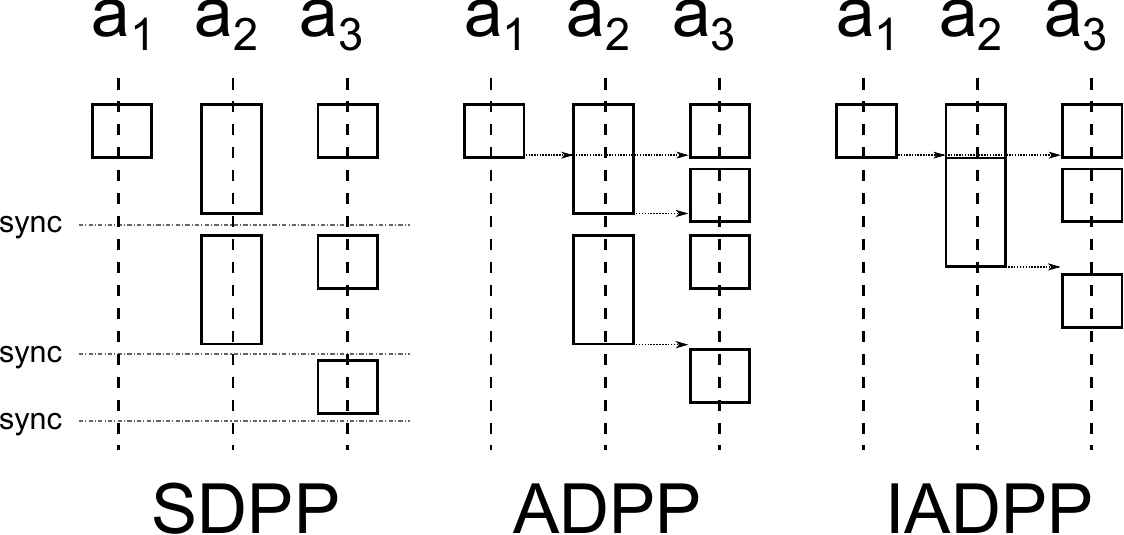}
\par\end{centering}

\caption{\label{fig:different_plan_lengths_execution_async.pdf}Sequence diagram
illustrating how can be wall-clock runtime further reduced by interrupting
the best-response planning. }
\end{figure}

The interruptible variant of ADPP strengthens the asynchronous aspect
of the ADPP. Figure~\ref{fig:different_plan_lengths_execution_async.pdf}
shows a another example of the decentralized algorithms execution
sequence. The total running time is 5 time units for SDPP and ADPP
while IADPP is able to shorten the execution to 4 time units. 

\subsection{Experimental evaluation}

We compare the centralized CA, SDPP, ADPP and IADPP on a few variants
of superconflict scenario and on a series of randomly generated problem
instances. The experiments were performed on Intel Core 2 Duo @ 2.1
Ghz. The problem instances used have the following common structure.
A given number of agents $n$ operate in a shared 20\,m x 20\,m
2-d square space. The agents generate a space-time trajectory between
their start and the destination position using a 4- or 8- connected
grid graph. The agents can move on the edges of the graph with the
constant speed of 1~m/s or they can wait for 0.5~s on any of the
vertices in the graph. The wait ``move'' can be used repeatedly.
The agents are required to maintain the separation distance 0.8~m
from all other agents at all times, even after they reached their
destination. 

The best-response planner used by all the agents is a spatio-temporal
A{*} planner operating over the grid graph, where the heuristic is
the time needed to travel the euclidean distance from the current
node to the destination node at the maximum speed. All the compared
algorithms use the identical best-response planner. 

To measure the runtime characteristics of the execution of decentralized
algorithms, we emulate the concurrent execution of the algorithms
using a discrete-event simulation. The simulation measures the execution
time of each message handling and uses the information to simulate
the concurrent execution of the decentralized algorithm as if it is
executed on $n$ independent computers. In the simulation we assume
zero communication delay. The concurrent process execution simulator
was implemented using Alite multi-agent simulation toolkit. The complete
source code of the experimental environment (including the concurrent
process simulator) and the video recordings of the experiments are
available at \href{http://agents.fel.cvut.cz/~cap/adpp/}{\url{http://agents.fel.cvut.cz/~cap/adpp/}}.

\subsection*{Superconflict scenarios}

\begin{figure}
\begin{centering}
\subfloat[\label{fig:Superconflict-scenario-example}Single superconflict scenario
example.]{\begin{centering}
\includegraphics[width=7cm]{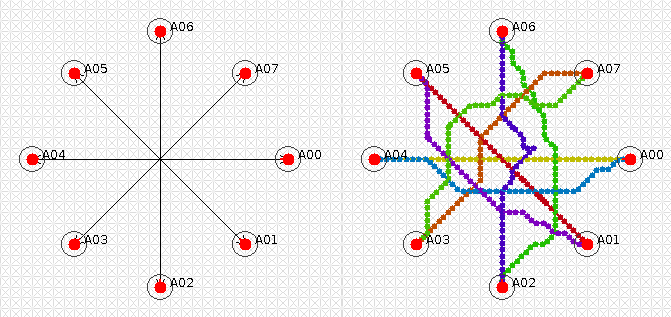}
\par\end{centering}

}
\par\end{centering}

\begin{centering}
\subfloat[\label{fig:Four-homogenous-superconflict}Four homogenous superconflicts
scenario example.]{\begin{centering}
\includegraphics[width=1\columnwidth]{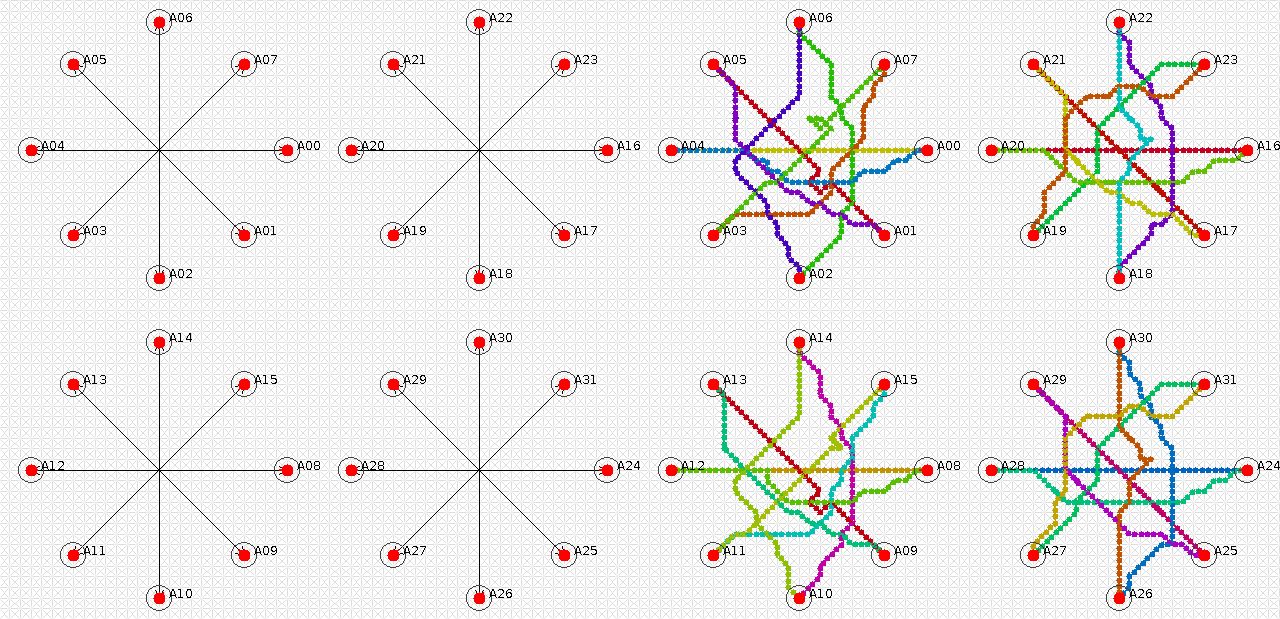}
\par\end{centering}

}
\par\end{centering}

\begin{centering}
\subfloat[\label{fig:Four-heterogeneous-superconflict}Four heterogeneous superconflicts
scenario example.]{\begin{centering}
\includegraphics[width=1\columnwidth]{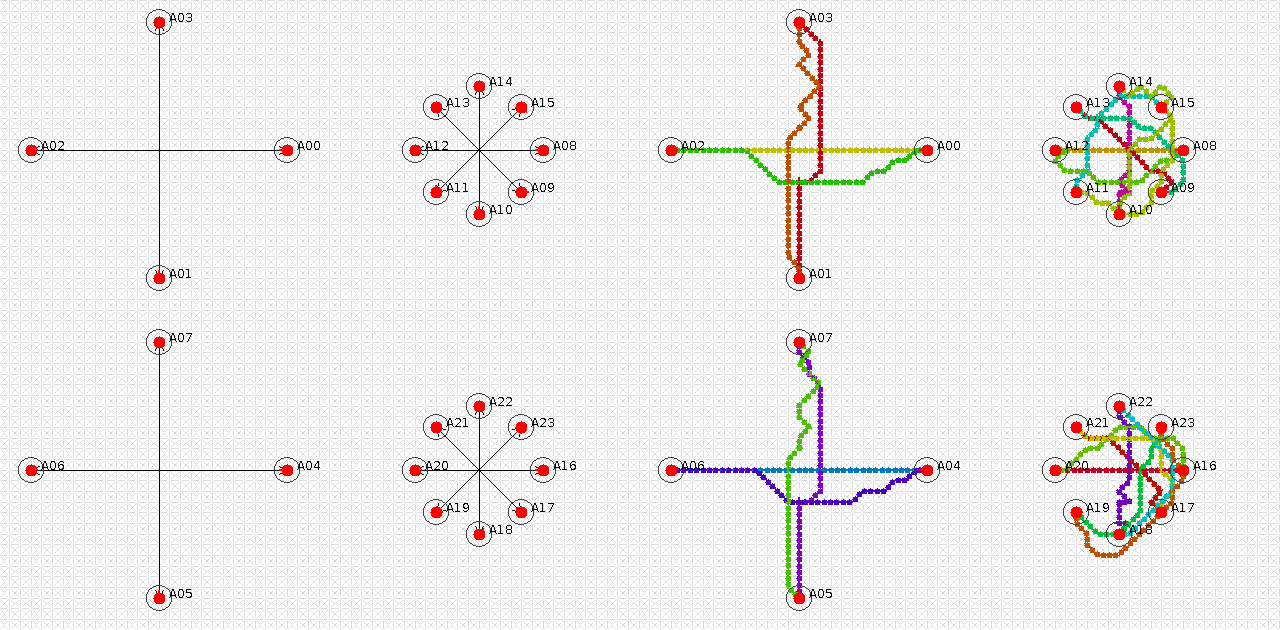}
\par\end{centering}

}
\par\end{centering}

\begin{centering}
\subfloat[\label{fig:Spiral-superconflict-scenario}Spiral superconflict scenario
example.]{\begin{centering}
\includegraphics[width=7cm]{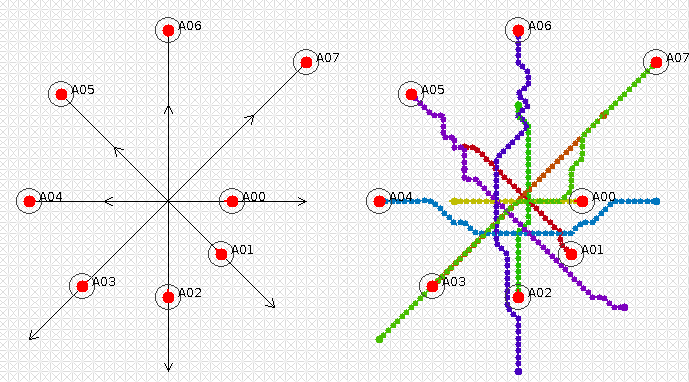}
\par\end{centering}

}
\par\end{centering}

\caption{Superconflict scenarios example -- problem configurations (left) and
solutions from IADPP algorithm (right).}
\end{figure}

We performed a number of experiments on a few variants of a challenging
superfconflict scenario. In the superconflict scenario, the agents'
start positions are put evenly spaced on a circle and their goal positions
are exactly at the opposite side of the circle. Therefore, the agents'
nominal trajectories all cross in the center of the circle. The superconflict
scenario is considered a challenging benchmark since each agent participating
in one superconflict circle is in conflict with all other agents of
that circle. Due to this coupling, the problem cannot be easily split
into independent subproblems and solved in parallel. In our implementation,
the agents plan their trajectory using a 60x60 8-connected grid graph.
We evaluated the algorithms on the following variants of superconflict
scenario:
\begin{description}
\item [{Single~supercoflict}] scenario with a 4\,meters-wide superconflict
of 8 agents placed in the middle of the square space. Agents' starting
configuration and the final trajectories obtained from IADPP are depicted
in Figure~\ref{fig:Superconflict-scenario-example}. Note that $A00$
is the highest priority agent in all our experiments.
\item [{Four~homogeneous~superconflicts}] scenario with four independent
superconflicts of 8 agents (4\,meters wide). This scenario allows
the cooperative pathfinding problem to be split into four independent
parts and thus the decentralized algorithms have an opportunity to
exploit the computational power of more processor (see Figure~\ref{fig:Four-homogenous-superconflict}). 
\item [{Four~heterogeneous~superconflicts}] scenario that combines two
superconflicts of four agents (4\,meters wide) and two superconflicts
of eight agents (only 2 meters wide). The former two have bigger radius
than the latter two and thus we expect that the best-response planner
invocations in the first group of superconflicts will take on average
longer to finish than the planners of the agents from the second group.
Such a difference in planning times leads to an inefficient execution
of SDPP, since the slowest progressing cluster of conflicts limits
the speed at which the other conflict clusters are resolved. The asynchronous
algorithm can resolve each of the superconflicts at a different pace
and thus we expect ADPP and IADPP to converge faster than SDPP (see
Figure~\ref{fig:Four-heterogeneous-superconflict}). 
\item [{Spiral~superconflict}] scenario is a superconflict of eight agents,
where the distance between an agent's start position and the center
of the superconflict increases with each agent. In our scenario the
radius varies between 2\,m and 6\,m. In result, the higher priority
agents often finish planning before the lower priority agents and
since all the agents are in mutual conflict, the planning process
of the lower priority agents is often invalidated. In both SDPP and
ADPP, the planning cannot be interrupted, and the agent will adapt
to the new situation only after the currently running planning process
finishes. Since the interruptible version of ADPP is designed to mitigate
this problem, we expect that it will outperform the other decentralized
methods in the scenario (see Figure~\ref{fig:Spiral-superconflict-scenario}). 
\end{description}
\begin{figure}
\begin{centering}
\includegraphics[width=3.1cm]{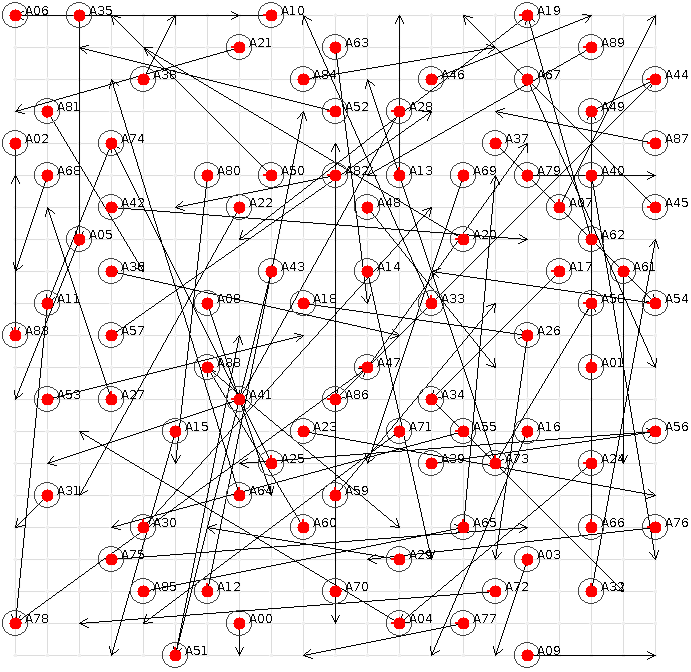}~~~\includegraphics[width=3.1cm]{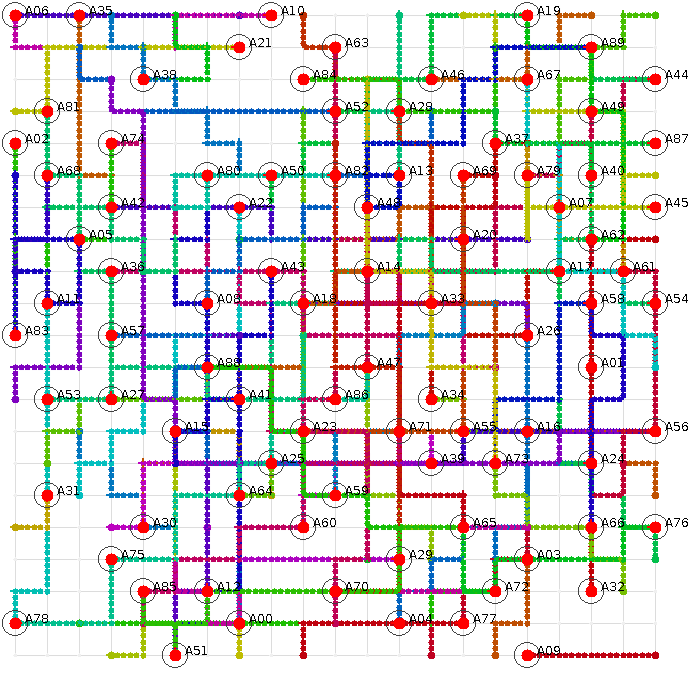}
\par\end{centering}

\centering{}\caption{\label{fig:random-scenario}One instance of random scenario with 90
agents. The start and goal position of each agent are depicted on
the left, the final solution found is on the right.}
\end{figure}

 Table~\ref{tab:Wallclock-runtimes-for} shows the wall-clock runtimes
of the four evaluated algorithms in the four presented scenarios.
For the single superconflict scenario, ADPP and IADPP runtimes are
close to CA, but SDPP shows significant synchronization overheads.
The second scenario in fact contains four independent instances of
the single superconflict as used in the first scenario. The total
complexity of this problem is expected to be four times higher than
that of the first scenario. The runtime of CA is more than quadrupled,
while the runtime of the decentralized algorithms stays almost unchanged,
which indicates perfect parallelization of the solution search process.
In the heterogeneous variant of the last scenario, the situations
looks different. As we can see from CA, the total complexity of the
problem is slightly lower than that of the first scenario. Due to
the differences in average planning times in the individual superconflicts,
the wall-clock runtime in SDPP is dominated by the slowest progressing
superconflict. We can see that both ADPP and IADPP can handle the
heterogeneity well. The spiral superconflict is a challenging scenario
for the non-interruptible asynchronous method. Thus, the ADPP wall-clock
runtime is closer to that of SDPP. 

\begin{table}
\begin{centering}
\begin{tabular}{|c|c|c|c|c|}
\hline 
 & CA & SDPP & ADPP & IADPP\tabularnewline
\hline 
\hline 
single superconflict & 10.30\,s & 26.24\,s & 11.91\,s & 9.50\,s\tabularnewline
\hline 
four homogeneous superconflicts & 45.81\,s & 26.97\,s & 13.86\,s & 11.62\,s\tabularnewline
\hline 
four heterogeneous superconflicts & 9.084\,s & 16.01\,s & 4.89\,s & 2.59\,s\tabularnewline
\hline 
spiral superconflict & 6.15\,s & 21.02\,s & 17.64\,s & 3.77\,s\tabularnewline
\hline 
\end{tabular}
\par\end{centering}

\caption{\label{tab:Wallclock-runtimes-for}Wall-clock runtimes for four versions
of superconflict scenario (averaged over 10 runs)}
\end{table}

\subsection*{Random scenario}

We measured the wall-clock runtime, communication complexity and solution
quality of the four algorithms CA, SDPP, ADPP and IADPP on a series
of problem instances that varied in the number of agents from 30 to
100. The start and goal vertices for each agent in the scenario were
selected randomly (see Figure~\ref{fig:random-scenario}). The distance
between the start and goal position was taken uniformly from the interval
$(5,10)$ and we further asserted that no two agents share the start
node and no two agents share the destination node. The agents plan
their trajectory on a 20x20 4-connected grid graph. For each number
of agents we ran 10 different random scenarios and averaged the results.
When any of the algorithms failed to find a solution to a problem
instance, the problem instance was excluded from the experiment. 

\begin{figure}
\subfloat[\label{fig:wallclock}Average wall-clock runtime for $n$-agent random
scenario]{\begin{centering}
\includegraphics[width=9cm]{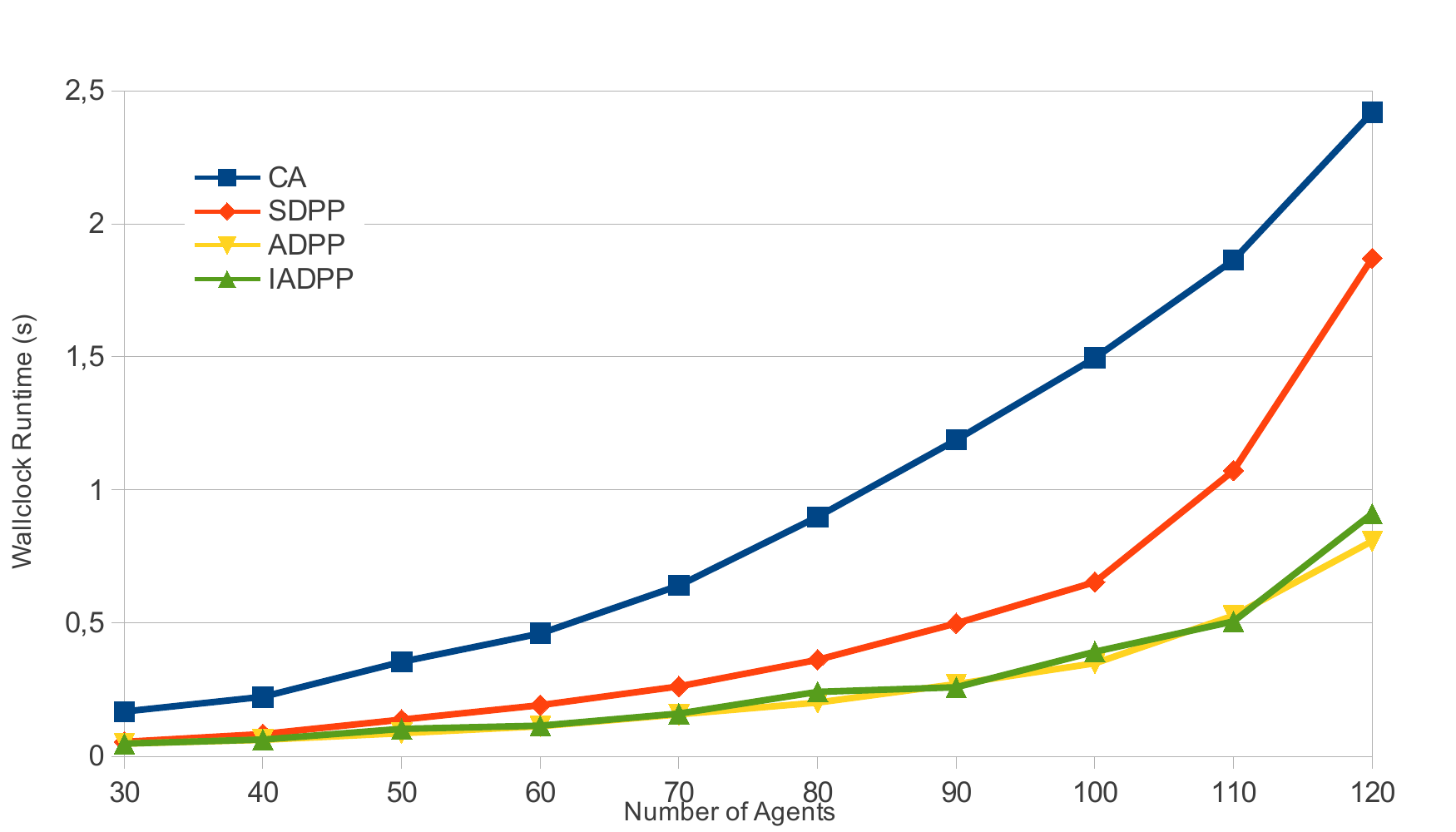}
\par\end{centering}

\centering{}}

\subfloat[\label{fig:messages}Average messages broadcasted]{\begin{centering}
\includegraphics[width=9cm]{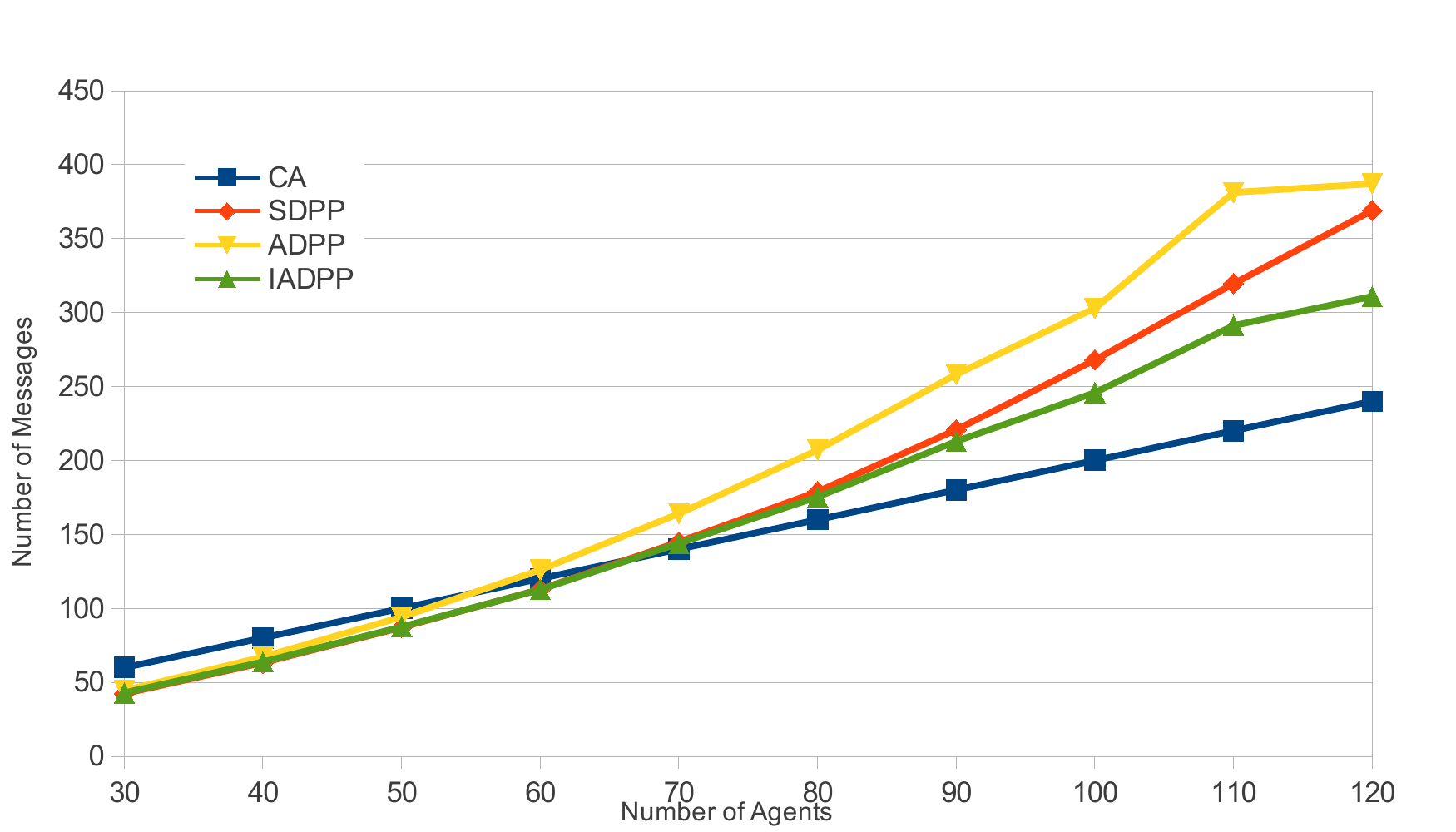}
\par\end{centering}

\centering{}}

\subfloat[\label{fig:cost}Average cost (prolongation of trajectories)]{\begin{centering}
\includegraphics[width=9cm]{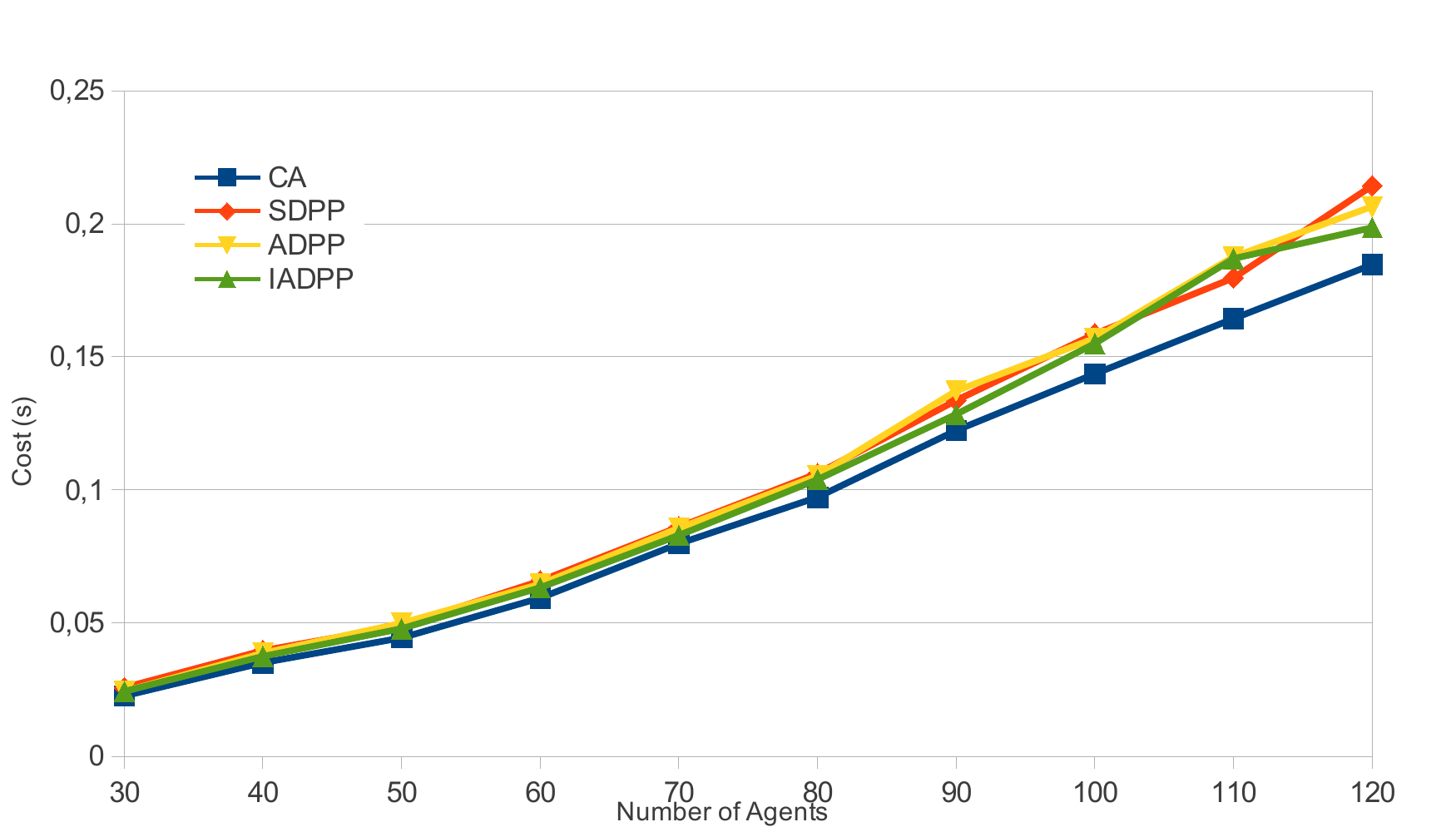}
\par\end{centering}

\centering{}}

\subfloat[\label{fig:failures}Failed instances ratio]{\begin{centering}
\includegraphics[width=9cm]{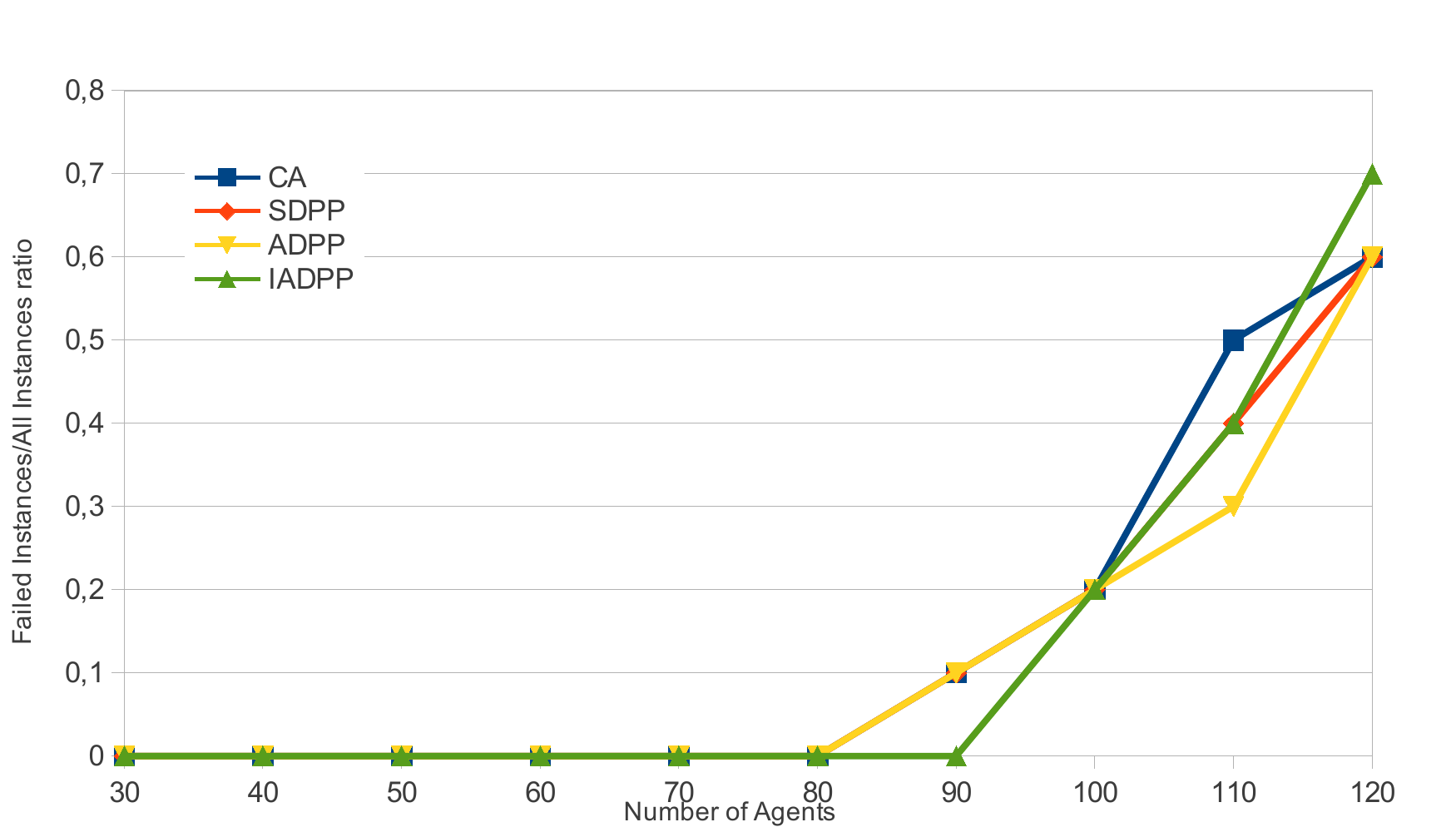}
\par\end{centering}

\centering{}}

\caption{Results from the random scenario}

\end{figure}

The wall-clock runtime represents the real-world time a particular
algorithm would need to converge to a solution. The wall-clock time
for CA is equal to its CPU-time and can be measured directly. The
average wall-clock runtime of the three decentralized algorithms on
random scenarios with $n$ agents was obtained by running an $n$
concurrent processes simulation of the algorithm execution. The results
for the wall-clock runtime experiment are shown in Figure~\ref{fig:wallclock}.
We can see that all decentralized algorithms can offer a speed-up
over the centralized solver. Further, we find that ADPP and IADPP
provide comparable wall-clock runtime performance, which is significantly
better than the runtime performance of SDPP, especially in dense problem
instances with many conflicting agents.

Further, we measured the communication complexity by counting the
messages each of the algorithms broadcasts during the execution. The
communication complexity of the CA algorithm is computed analytically.
We assume that the algorithm is used to coordinate paths in a distributed
system in the following way. All the agents are required to communicate
their objectives to the central solver. When the central solver finishes
the planning, it informs each agent about its new path. Thus, we use
$2n$ as the communication complexity of the centralized solver. In
Figure~\ref{fig:messages} we can see that the decentralized algorithms
start exceeding the communication complexity of the centralized solution
for scenarios with more than 60 agents. Further, we find that IADPP
algorithm has lower communication complexity than ADPP. This can be
explained by looking at how the two algorithms react to an inform
message that invalidates the current running planning effort. In ADPP,
the planning is finished, the new plan broadcast and only after that
a new planning is started. In IADPP, the planner is restarted quietly,
yielding no extra communication.

Figure~\ref{fig:cost} shows the quality of the generated solutions.
The reason why decentralized algorithms return on average slightly
worse solutions than the CA algorithm lies in the replanning condition
used by the decentralized algorithms. The condition states that an
agent should replan his trajectory only if the trajectory is inconsistent
with his agentview. In result, the agent may receive an updated trajectory
from a higher-priority agent that allows for improvement in his current
trajectory, but since the trajectory may be still consistent, the
agent will not exploit such an improvement opportunity. 

Finally, Figure~\ref{fig:failures} shows the failure rates of the
individual algorithms as a function of the number of agents in a scenario.

\subsubsection*{}

\section{Conclusion \label{sec:Conclusion}}

In this paper we introduced an asynchronous decentralized prioritized
planning algorithm for space-time cooperative pathfinding problem.
Two variants of the algorithm, ADPP and IADPP, were presented. We
proved the correctness and termination of both introduced algorithms.
The algorithms were compared to both central and decentralized state-of-the-art
techniques for prioritized planning. Experimental validation and evaluation
showed the benefits and limitations of the discussed algorithms. The
experiments show the advantages of asynchronous and interruptible
execution of the presented algorithms on a set of superconflict scenarios. 

The large scale evaluation on a set of random problem instances documents
a significant reduction of average wall-clock runtime of both ADPP
and IADPP in comparison to the centralized (approx. 65\% time reduction)
and the decentralized synchronous algorithm (approx. 45\% time reduction).
The communication complexity is the worst for ADPP, while IADPP is
still better than SDPP, but worse than CA for higher numbers of agents.
The average cost of generated solutions is similar for all decentralized
algorithms and only approx. 10\% worse than CA. The failure ratio
of all prioritized methods is comparable. The experimental validation
fully supports the expectations on the improvements of the ADPP and
IADPP over both CA and SDPP.

\subparagraph*{Acknowledgements}

This work was supported by the Ministry of Education, Youth and Sports
of Czech Republic within the grant no. LD12044.

\bibliographystyle{plain}
\bibliography{../../bib}

\end{document}